\PassOptionsToPackage{hyphens}{url}
\PassOptionsToPackage{dvipsnames,svgnames,x11names}{xcolor}
\documentclass[
  12pt]{article}

\usepackage{amsmath,amssymb}

\usepackage[T1]{fontenc}
\usepackage[utf8]{inputenc}
\usepackage{textcomp}

\usepackage{lmodern}

\IfFileExists{upquote.sty}{\usepackage{upquote}}{}
\IfFileExists{microtype.sty}{
  \usepackage[]{microtype}
  \UseMicrotypeSet[protrusion]{basicmath} 
}{}

\makeatother
\usepackage{xcolor}
\makeatletter
\ifx\paragraph\undefined\else
  \let\oldparagraph\paragraph
  \renewcommand{\paragraph}{
    \@ifstar
      \xxxParagraphStar
      \xxxParagraphNoStar
  }
  \newcommand{\xxxParagraphStar}[1]{\oldparagraph*{#1}\mbox{}}
  \newcommand{\xxxParagraphNoStar}[1]{\oldparagraph{#1}\mbox{}}
\fi
\ifx\subparagraph\undefined\else
  \let\oldsubparagraph\subparagraph
  \renewcommand{\subparagraph}{
    \@ifstar
      \xxxSubParagraphStar
      \xxxSubParagraphNoStar
  }
  \newcommand{\xxxSubParagraphStar}[1]{\oldsubparagraph*{#1}\mbox{}}
  \newcommand{\xxxSubParagraphNoStar}[1]{\oldsubparagraph{#1}\mbox{}}
\fi
\makeatother

\usepackage{longtable,booktabs,array}
\usepackage{calc} 
\usepackage{etoolbox}
\makeatletter
\patchcmd\longtable{\par}{\if@noskipsec\mbox{}\fi\par}{}{}
\makeatother
\IfFileExists{footnotehyper.sty}{\usepackage{footnotehyper}}{\usepackage{footnote}}
\makesavenoteenv{longtable}
\usepackage{graphicx}
\makeatletter
\def\maxwidth{\ifdim\Gin@nat@width>\linewidth\linewidth\else\Gin@nat@width\fi}
\def\maxheight{\ifdim\Gin@nat@height>\textheight\textheight\else\Gin@nat@height\fi}
\makeatother
\setkeys{Gin}{width=\maxwidth,height=\maxheight,keepaspectratio}
\makeatletter
\def\fps@figure{htbp}
\makeatother

\makeatletter
\@ifpackageloaded{caption}{}{\usepackage{caption}}
\AtBeginDocument{%
\ifdefined\contentsname
  \renewcommand*\contentsname{Table of contents}
\else
  \newcommand\contentsname{Table of contents}
\fi
\ifdefined\listfigurename
  \renewcommand*\listfigurename{List of Figures}
\else
  \newcommand\listfigurename{List of Figures}
\fi
\ifdefined\listtablename
  \renewcommand*\listtablename{List of Tables}
\else
  \newcommand\listtablename{List of Tables}
\fi
\ifdefined\figurename
  \renewcommand*\figurename{Figure}
\else
  \newcommand\figurename{Figure}
\fi
\ifdefined\tablename
  \renewcommand*\tablename{Table}
\else
  \newcommand\tablename{Table}
\fi
}
\@ifpackageloaded{float}{}{\usepackage{float}}
\floatstyle{ruled}
\@ifundefined{c@chapter}{\newfloat{codelisting}{h}{lop}}{\newfloat{codelisting}{h}{lop}[chapter]}
\floatname{codelisting}{Listing}

\makeatother

\usepackage{natbib}
\usepackage{bookmark}

\IfFileExists{xurl.sty}{\usepackage{xurl}}{} 
\hypersetup{
  pdftitle={Time-Aware Attention for Enhanced Electronic Health Records Modeling},
  pdfauthor={Junhan Yu; Zhunyi Feng; Junwei Lu; Tianxi Cai; Doudou Zhou},
  colorlinks=true,
  linkcolor={blue},
  filecolor={Maroon},
  citecolor={Blue},
  urlcolor={Blue},
  pdfcreator={LaTeX via pandoc}}
\usepackage{hyperref}
\usepackage{mathtools}
\usepackage{amsfonts}
\usepackage{algorithm}
\usepackage[noend]{algpseudocode}
\usepackage{wrapfig}
\usepackage{graphicx}
\usepackage{caption}
\usepackage{xcolor}
\usepackage{amsmath}
\usepackage{enumerate}
\usepackage{amsthm}
\usepackage{multirow}
\usepackage{url}
\usepackage{booktabs}
\usepackage{nicefrac}
\usepackage{xcolor}
\usepackage{amsmath}
\usepackage{amssymb}
\usepackage{amsfonts}
\usepackage[capitalize,noabbrev]{cleveref}

\def\spacingset#1{\renewcommand{\baselinestretch}%
{#1}\small\normalsize} \spacingset{1}

\newcount\comments  
\newcommand{\genComment}[2]{\ifnum\comments=1{\textcolor{#1}{\textsf{\footnotesize #2}}}\fi}

\def\spacingset#1{\renewcommand{\baselinestretch}%
{#1}\small\normalsize} \spacingset{1}

\newcommand{\anon}{1}

\begin{document}

\if1\anon
{
  \title{\bf Time-Aware Attention for Enhanced Electronic Health Records Modeling}
    \author{
    Junhan Yu\textsuperscript{1}\thanks{Equal contribution.},
    Zhunyi Feng\textsuperscript{2}\footnotemark[1], 
    Junwei Lu\textsuperscript{3},  \\
    Tianxi Cai\textsuperscript{3}, 
    Doudou Zhou\textsuperscript{1}\thanks{Corresponding author.} \bigskip \\
    \small 
    \textsuperscript{1} Department of Statistics and Data Science, National University of Singapore  \\
    \small 
    \textsuperscript{2} School of Computing, National University of Singapore  \\
    \small 
    \textsuperscript{3} Department of Biostatistics, Harvard T.H. Chan School of Public Health, USA\\
    \small 
    }
    \date{}
  \maketitle
} \fi

\if0\anon
{
  \bigskip
  \bigskip
  \bigskip
  \begin{center}
    {\LARGE\bf Time-Aware Attention for Enhanced Electronic Health Records Modeling}
\end{center}
  \medskip
} \fi

\bigskip
\begin{abstract}
Electronic Health Records (EHR) contain valuable clinical information for predicting patient outcomes and guiding healthcare decisions.  However, effectively modeling Electronic Health Records (EHRs) requires addressing data heterogeneity and complex temporal patterns. Standard approaches often struggle with irregular time intervals between clinical events. We propose TALE-EHR, a Transformer-based framework featuring a novel time-aware attention mechanism that explicitly models continuous temporal gaps to capture fine-grained sequence dynamics. To complement this temporal modeling with robust semantics, TALE-EHR leverages embeddings derived from standardized code descriptions using a pre-trained Large Language Model (LLM), providing a strong foundation for understanding clinical concepts. Experiments on the MIMIC-IV and PIC dataset demonstrate that our approach outperforms state-of-the-art baselines on tasks such as disease progression forecasting. TALE-EHR underscores the benefit of integrating explicit, continuous temporal modeling with strong semantic representations  provides a powerful solution for advancing EHR analysis.
\end{abstract}

\section{Introduction}
Deep learning approaches, particularly transformer-based architectures, have propelled significant advancements in natural language processing (NLP). Within healthcare, Electronic Health Records (EHRs) offer a rich repository of information, encompassing patient medical histories, diagnoses, laboratory results, and clinical notes. However, harnessing this data for predictive modeling remains challenging due to its fragmented, sensitive, and heterogeneous nature, alongside its crucial inherent temporal complexity. Accurately capturing the sequence, timing, and duration between clinical events is paramount for understanding patient trajectories and predicting future outcomes.

Modeling the temporal structure of clinical events is essential. The sequence and precise timing of diagnoses, medications, and interventions carry vital information. Short intervals might indicate acute episodes, whereas longer intervals could reflect chronic condition management or latency periods. To address this critical temporal dimension, we introduce a novel time-aware attention mechanism. This mechanism moves beyond standard self-attention by explicitly incorporating the continuous temporal intervals between clinical events into the attention score calculation. This enables the Transformer model to dynamically weight the influence of past events based not only on their semantic similarity but crucially on their temporal proximity, thereby capturing both semantic content and the fine-grained temporal dynamics inherent in patient data.

Furthermore, Conventional EHR code embeddings often struggle with semantic nuances. LLMs pre-trained on textual code descriptions \citep{hur2022unifyingheterogeneouselectronichealth, wang2022transtablearningtransferabletabular, wang2024meditabscalingmedicaltabular, su2025multimodalmedicalcodetokenizer} offer richer semantics, overcoming heterogeneity. We leverage this for robust code embeddings, enhancing performance even with limited data.

To integrate these insights, we propose the Time-Aware Language Encoder for EHR (TALE-EHR), a transformer-based point process framework. TALE-EHR’s core innovation lies in its time-aware attention mechanism, specifically designed for the irregularities of EHR event sequences. This is complemented by leveraging embeddings derived from clinical code descriptions via a pre-trained LLM, ensuring robust semantic understanding and facilitating better generalization across healthcare institutions with diverse coding practices. By processing the textual descriptions for 12,232 distinct medical codes within a temporally sophisticated architecture, TALE-EHR provides a comprehensive and interpretable representation of patient trajectories, leading to improved predictive performance.

Our key contributions are as follows. First, we design a novel time-aware attention mechanism that explicitly models continuous temporal intervals between clinical events within a Transformer framework, improving the capture of time-dependent relationships in EHR data. Second, we leverage pre-trained LLM embeddings derived from standardized code descriptions. This provides our model with strong initial semantic understanding, which, when combined with our temporal modeling, leads to enhanced representation robustness and improved predictive performance, particularly in handling diverse clinical codes. Third, our framework uniquely combines enhanced semantic understanding with explicit temporal dynamic modeling, offering a more holistic approach to EHR sequence analysis. 

To empirically validate TALE-EHR's effectiveness and generalizability, we conducted extensive experiments on two distinct EHR datasets: the large-scale MIMIC-IV database \citep{johnson2023mimic} (over 360,000 adult patient records) and the Paediatric Intensive Care (PIC) database \citep{Zeng2020}. On MIMIC-IV, TALE-EHR achieves state-of-the-art performance in predicting next-visit medical events across all 12,232 unique codes in our processed dataset. Furthermore, it demonstrates robust prediction for specific diseases on MIMIC-IV, attaining an average Area Under the Curve (AUC) of 0.926 across these evaluated conditions. Complementary experiments on the PIC dataset further underscore our method's effectiveness and robustness across different clinical settings and patient populations. These comprehensive evaluations highlight TALE-EHR's capabilities, particularly its enhanced temporal modeling, across a variety of clinical prediction tasks.

\section{Related Work}
The representation of EHR for predictive modeling has been extensively studied, with approaches showing diverse strategies for capturing and utilizing the complex information within EHR data, particularly its temporal dynamics.

Early works employed RNNs and variants \citep{choi2016learning, du2016recurrent, shang2021interpretable} for sequential dependencies but faced limitations with long-range dependencies and temporal irregularities common in EHRs. Knowledge graph-guided \citep{zhang2019, wang2020, li2019} and contrastive learning \citep{shang2019event} approaches have also been explored, yet often rely on curated structures or prioritize structured codes over nuanced temporal dynamics \citep{chen2020modeling}.
Transformer-based models \citep{lee2020, choi2020, lample2019} excel at long-range dependencies, but effectively integrating temporal information specific to EHRs remains an active research area.

Recognizing temporal importance, studies have incorporated various strategies. Some focused on sequence order or semantic-based attention within RNNs \citep{choi2017retaininterpretablepredictivemodel, Ma_2017}, or combined RNNs with point process theory or time-aware gates \citep{du2016recurrent, liu2019point, Liu2022-pv}. Within Transformer architectures, approaches include hierarchical time-aware attention \citep{Luo2020}, discrete time tokens \citep{pang2021cehrbertincorporatingtemporalinformation}, standard time embeddings with specialized output layers \citep{steinberg2024motor}, text serialization relying on positional embeddings \citep{lee2024emergencydepartmentdecisionsupport}, or retrieval-enhanced methods using event timestamps alongside embeddings \citep{kim2024generalpurposeretrievalenhancedmedicalprediction}.

While advancing the field, these methods often handle temporal information indirectly, e.g., via sequence order, separate time embeddings, discrete markers, or fixed structural integrations, potentially limiting direct, adaptive modulation by continuous temporal differences.

In contrast, TALE-EHR introduces a unique time-aware attention mechanism by directly incorporating a learnable temporal weighting function, $w(t)$, into the Transformer's self-attention computation. This function dynamically adjusts attention scores based on the continuous time difference ($|t_j - t_k|$) between events. Unlike approaches relying primarily on sequence order, pre-fusing time embeddings, or using discrete markers, TALE-EHR's $w(t)$ directly modulates semantic relevance by continuous time differences at the very moment of attention calculation. This allows for adaptive capture of complex temporal dependencies, promoting a deep fusion of semantic and temporal information for richer representations, rather than treating time as a preprocessing step or an auxiliary feature.
\section{Method}
\label{sec:method}

EHR data contain rich temporal information that reflects patients' clinical trajectories. We model this data as a \textit{marked temporal point process}, where each patient $i$ is represented by a sequence of time-stamped clinical events:
$$\mathcal{H}^{(i)} = \{(t_1^{(i)}, c_1^{(i)}), (t_2^{(i)}, c_2^{(i)}), \dots, (t_{m_i}^{(i)}, c_{m_i}^{(i)})\}$$
with $t_1^{(i)} \leq t_2^{(i)} \leq \ldots t_{m_i}^{(i)}$. Here, $t_j^{(i)}$ represents the timestamp of the $j$-th clinical event, $c_j^{(i)} \in \mathcal{C}$ with $\mathcal{C}$ being the set of all possible EHR codes, and $m_i$ is the total number of clinical events in the patient $i$.

Following the point process framework \citep{daley2003introduction}, let $N^{(i)}$ denote the counting process for patient $i$, defined by occurrence times $\{t_1^{(i)}, t_2^{(i)}, ...\}$. For a time interval $A \subseteq \mathbb{R}^+$, we define $N^{(i)}(A) = \big| \{t_j^{(i)}: t_j^{(i)} \in A\} \big|$ as the number of events in $A$. The patient history up to time $t$ is denoted as $\mathcal{H}_t^{(i)} = \{ (t_j^{(i)}, c_j^{(i)}): t_j^{(i)} \leq t\}$.

Additionally, each patient is assumed to have a demographic covariate vector $\mathbf{Z}_i(t) \in \mathbb{R}^p$, which includes both time-invariant covariates (e.g., sex and race) and time-varying covariates (e.g., age). Based on this, the conditional intensity function is formally defined as:
\begin{equation*}
    \lambda^{(i)}(t) = \lim_{{\rm d} t \rightarrow 0} \mathbb{E}[N^{(i)}({\rm d} t)|\mathcal{H}_t^{(i)},\mathbf{Z}_i(t)]/{\rm d} t
\end{equation*}
where $N^{(i)}({\rm d} t) = N^{(i)}\big([t, t + {\rm d} t) \big)$ denotes the number of events occurring in the infinitesimal interval $[t, t + {\rm d} t)$. The conditional intensity function $\lambda^{(i)}(t)$ describes the instantaneous rate at which events are expected to occur at time $t$, given the patient's history $\mathcal{H}_t^{(i)}$ and covariates $\mathbf{Z}_i(t)$.

To operationalize this theoretical framework with modern deep learning, we implement the conditional intensity function using transformer-based architectures. Our model aims to achieve two key objectives: (1) Event Timing Prediction: the timing of future events is modeled by a parameterized intensity function:
    \begin{equation}
        \lambda^{(i)}(t) = g(\mathbf{Z}_i(t), \mathbf{h}_t^{(i)}); 
        \label{eq:g}
    \end{equation}
and (2) Event Prediction: the type of the next clinical event is predicted via:
    \begin{equation}
        P( c_t \mid \mathbf{Z}_i(t), \mathcal{H}_t^{(i)}) = f(\mathbf{Z}_i(t), \mathbf{h}_t^{(i)}).
        \label{eq:f}
    \end{equation}
Here, $c_t$ is the medical code for the next event after time $t$,  $\mathbf{h}_t^{(i)}$ is a learned representation of patient history $\mathcal{H}_t^{(i)}$, and $g$ and $f$ are neural network functions.

\begin{figure}[t]
    \centering
    \includegraphics[width=0.85\linewidth]{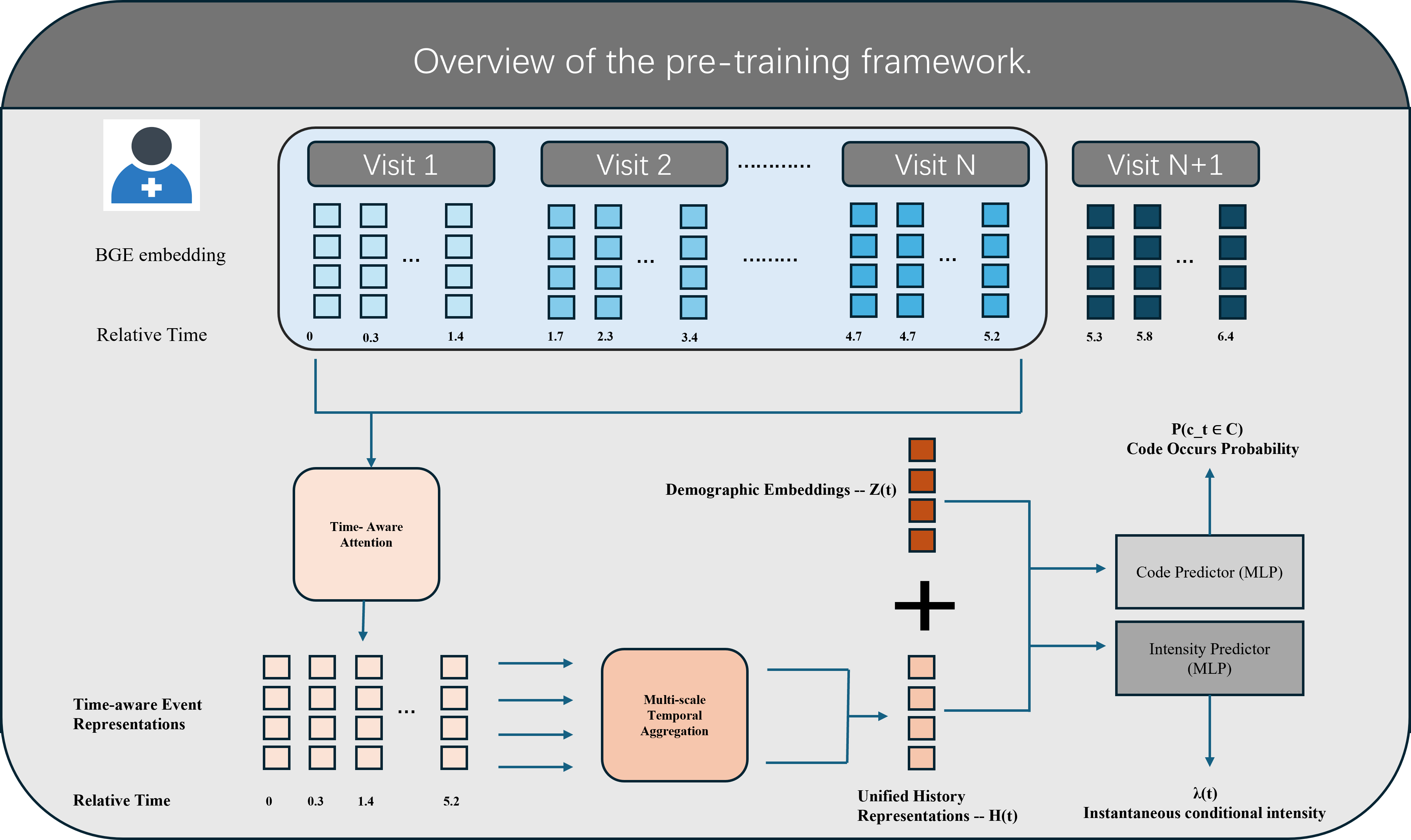}
   \caption{Overview of the TALE-EHR framework. }
    \label{fig:pretrain}
\end{figure}

The proposed TALE-EHR algorithm, as illustrated in Figure ~\ref{fig:pretrain}, leverages three core innovations to effectively model clinical event sequences. First, we introduce a time-aware attention mechanism that explicitly captures temporal relationships between events, enabling the model to effectively account for complex dependencies in patient histories (Section~\ref{sec:time_attention}). Second, the framework incorporates multi-scale temporal features to handle patterns occurring across varying time scales, ensuring robust representation of both short-term and long-term dynamics (Section~\ref{sec:multi_scale}). Third, medical codes are embedded using pre-trained language models, providing a unified and interpretable representation of clinical events.We propose a joint training strategy that optimizes both timing and content prediction objectives (Section~\ref{sec:training_objective}). This allows the model to generalize across a wide range of downstream clinical tasks (Section \ref{sec:fine_tuning}).

\subsection{Time-aware Attention Mechanism}
\label{sec:time_attention}

A core component of our approach is the time-aware attention mechanism, which explicitly captures temporal relationships between clinical events. Given the history $\mathcal{H}_t$ (superscript $(i)$ omitted for simplicity), each event is encoded through the attention mechanism as: 
\begin{equation}
   E_{c_j} (t) = \sum_{(c_k,t_k) \in \mathcal{H}_t} A(Q_{c_j}, K_{c_k}, |t_j - t_k| ) V_{c_k}, 
\end{equation}
where $(c_j, t_j) \in \mathcal{H}_t$. To generate robust semantic embeddings ($\mathbf{v}_c$) for medical codes from their textual descriptions, we utilize BGE~\citep{chen2024bge}, a powerful general-purpose text encoder pre-trained on diverse, broad-domain corpora. This approach helps mitigate potential biases from medical-specific pre-training and enhances generalization. The query ($Q_c$), key ($K_c$), and value ($V_c$) vectors are then derived by applying separate multi-layer perceptrons (MLPs) to these pre-trained LLM embeddings $\mathbf{v}_c$, i.e., $Q_c = \text{MLP}_Q(\mathbf{v}_c)$, $K_c = \text{MLP}_K(\mathbf{v}_c)$, and $V_c = \text{MLP}_V(\mathbf{v}_c)$. This allows the model to learn task-specific projections from the rich semantic information encoded in $\mathbf{v}_c$ while maintaining computational efficiency by keeping the underlying BGE embeddings fixed during training.

The time-aware attention function $A(\cdot)$ captures both semantic relationships and temporal dependencies:
\begin{equation}
   A(Q,K, t) = \text{softmax}\left(\frac{Q^\top K}{\sqrt{d}}w( t)\right)
\end{equation}
where $w( t)$ is a learnable temporal weighting function
\begin{equation*}
    w( t) = \sigma(a_0 + a_1t + a_2t^2 + ... + a_s t^s)
\end{equation*}
Here, $\{a_0, \dots, a_s\}$ are learnable coefficients for $w(t)$, and $\sigma$ is a sigmoid ensuring temporal weights are within $[0, 1]$ for stability. The polynomial design of $w(t)$ offers a flexible and interpretable framework for modeling temporal influence. Our ablation studies (Section~\ref{sec:ablation}) confirm that the polynomial basis (Order 5) provides a strong balance of expressiveness, structured modeling, and efficiency for nuanced temporal dependency capture, outperforming simpler (piecewise) or comparably performing but more computational cost (MLP) alternatives.

This time-aware attention mechanism effectively captures both semantic relationships and temporal dependencies among clinical events. By dynamically adjusting influence based on temporal proximity, it establishes a robust foundation for handling complex temporal patterns in patient histories.

\subsection{Multi-scale Temporal Representation}
\label{sec:multi_scale}

Clinical events exhibit intricate temporal dependencies across varying time scales. For example, a patient's current condition might be influenced by both recent medications and chronic diseases diagnosed years ago. Furthermore, the irregularity in the timing and number of historical events across patients introduces additional challenges. To address these complexities, we construct a unified representation of patient history, $\mathbf{h}_t^{(i)}$, using a hierarchical attention mechanism: 
\begin{equation*}
    \mathbf{h}_t^{(i)} = \sum_{j=1}^m \alpha_j (t) E_{c_j}(t),
\end{equation*}
where $m = |\mathcal{H}_t^{(i)}|$ is the number of events before time $t$ for patient $i$.

The attention weights $\alpha_j(t)$ incorporate both semantic relevance and temporal proximity:
\begin{equation*}
    \alpha_j(t) = \frac{\exp((Q_{\text{base}}^\top E_{c_j}(t))w(|t-t_j|))}{\sum_{k=1}^m \exp((Q_{\text{base}}^\top E_{c_k}(t))w(|t-t_k|))}.
\end{equation*}
Here, $Q_{\text{base}}$ is a learnable query vector optimized to identify historically relevant events, and the learned temporal weighting function $w(|t - t_{k,j}|)$ adjusts event influence based on temporal distance. This allows the model to prioritize clinically significant events while modulating their relevance by time, with softmax normalization enabling focus on multiple relevant events simultaneously.

This temporal attention mechanism provides several critical capabilities: it captures both fine-grained intra-visit and long-term inter-visit temporal patterns (e.g., sequential ordering, chronic disease progression), effectively handles irregular temporal intervals common in clinical data, learns disease progression dynamics across diverse time scales, and models complex interactions among co-occurring medical conditions, including simultaneous and temporally dependent events.

Through this mechanism, the unified representation $\mathbf{h}_t^{(i)}$ encodes a patient's comprehensive health state at time $t$, capturing both temporal patterns and clinical relationships. Beyond predicting visit timing and codes, $\mathbf{h}_t^{(i)}$ can be leveraged for various downstream clinical predictions, including disease onset and other health outcomes.

\subsection{Joint Training Objective}
\label{sec:training_objective}

The model is trained to predict both the timing and medical codes of future visits through \eqref{eq:g} and \eqref{eq:f}, where functions $g$ and $f$ are implemented as multilayer perceptrons with GELU activation. Specifically, both networks first project demographic features and history representations into lower-dimensional spaces independently through single hidden layers, then concatenate and process them through fully connected layers, a 4-layer network for $g$ to output the intensity value and a 3-layer network for $f$ to predict code probabilities.

\paragraph{Temporal Point Process Loss.} 
To model visit time, we minimize a least-squares functional derived from empirical risk minimization principles \citep{geer2000empirical, massart2007concentration, koltchinskii2011oracle, Bartlett2006}:
\begin{equation*}
    \mathcal{L}_{\text{time}}(\theta) = \|\lambda_\theta\|^2_T - \frac{2}{T}\int_{[0,T]} \lambda_\theta(t) dN(t),
\end{equation*}
where $\|\lambda_\theta\|^2_T $ denotes the squared norm of the intensity function, and the second term measures the model's fit at event times. By the Doob-Meyer decomposition, this loss is minimized when $\lambda_\theta$ converges to the true intensity function.

Using Monte Carlo sampling, the squared norm term is approximated as: 
\begin{equation*}
    \|\lambda_\theta\|^2_T = \frac{1}{T} \int_{[0,T]} \lambda_\theta(t)^2 dt \approx \frac{1}{N} \sum_{i=1}^N \lambda_\theta(t_i)^2,
\end{equation*}
where $t_i$ are time points  uniformly sampled from the interval $[0,T]$, and $N$ is the number of samples used to approximate the integral. The second term is computed using the observed event times: 
\begin{equation*}
    \frac{2}{T} \int_{[0,T]} \lambda_\theta(t) dN(t) = \frac{2}{T} \sum_{ (t_j,c_j) \in \mathcal{H}_t } \lambda_\theta(t_j).
\end{equation*}

\paragraph{Clinical Code Prediction Loss.} To address the inherent class imbalance in clinical code prediction, we adopt a focal loss:
\begin{equation*}
\mathcal{L}_{\text{code}} = -\sum_{c \in \mathcal{C}} \alpha(1-p_c)^\gamma \tilde{y}_c\log(p_c),
\end{equation*}
where $p_c$ is the predicted probability for code $c$, and $\tilde{y}_c$ is the smoothed ground truth label ($0.95$ for positive cases and $0.05$ for negative cases). The parameters $\alpha=0.25$ and $\gamma=2.0$ are standard choices that balance the loss contributions and emphasize harder examples. This dynamic adjustment ensures that the model focuses on rare, misclassified examples.

The combined loss function for joint optimization is $\mathcal{L} = \mathcal{L}_{\text{time}} + \gamma \mathcal{L}_{\text{code}}$, 
where $\gamma$ is a hyperparameter balancing the temporal and clinical prediction tasks. This joint optimization framework enables the model to learn both temporal patterns and clinical relationships, providing a holistic understanding of a patient’s healthcare trajectory.

\subsection{Fine-tuning for Downstream Tasks}
\label{sec:fine_tuning}

After pre-training our model on the temporal point process objectives, we fine-tune it for downstream clinical prediction tasks including disease onset prediction. For each task $k$, the model processes 1024 clinical events before the target event (e.g., the first occurrence of a disease) to generate the corresponding history representation $\mathbf{h}_{t}^{(i)}$ and predict the outcome through task-specific MLPs: $p_k^{(i)} = \text{MLP}_k(\mathbf{h}_{t}^{(i)})$, 
where $p_k^{(i)}$ represents the predicted probability for task $k$. The model is trained using weighted binary cross-entropy loss to handle class imbalance:
\begin{equation*}
    \mathcal{L}_{\text{task}} = -\sum_{i=1}^N w_k [y_k^{(i)}\log(p_k^{(i)}) + (1-y_k^{(i)})\log(1-p_k^{(i)})],
\end{equation*}
where $w_k$ is the task-specific weight. To preserve the temporal patterns learned during pre-training, we employ a differential learning rate strategy, applying a lower learning rate for pre-trained parameters and a higher rate for task-specific layers. Finally, we summarize the procedure in Algorithm \ref{app:training} in Appendix \ref{app:algorithm}. 

\section{Experiments}
\label{sec:experiment}

\subsection{Dataset and Preprocessing}
We evaluate TALE-EHR on two distinct EHR datasets: the large-scale adult MIMIC-IV database \citep{johnson2023mimic} (inpatient, ICU, emergency visits; including diagnoses, medications, procedures, labs) and the Paediatric Intensive Care (PIC) database \citep{Zeng2020} (paediatric ICU data; similar clinical information extracted).

Our preprocessing pipeline standardizes both datasets. A roll-up strategy aggregates lower-level medical codes into higher-level categories (diagnoses, medications, procedures, labs) to reduce sparsity and improve comparability, ensuring appropriate granularity (MIMIC-IV roll-up details in Appendix \ref{app:roll-up}). 
For temporal encoding, event timestamps are converted to relative distances from the patient's initial record ($t=0$), normalized to a consistent unit (e.g., weeks), and then logarithmically transformed to handle large variations.

The processed MIMIC-IV dataset contains 12,232 unique codes from 342,917 patient records; the PIC dataset yields 2,607 codes from 12,868 records. Both datasets are partitioned into 70\% training, 10\% validation, and 20\% testing sets.

\subsection{Training Procedure}
Training is two-stage. Pre-training (Section~\ref{sec:training_objective}) learns temporal point process objectives for $10$ epochs (Adam, LR $10^{-4}$, batch $16$). Fine-tuning (Section~\ref{sec:fine_tuning}) then adapts the model for disease and clinical benchmark prediction. During fine-tuning, most pre-trained parameters are frozen, except for the trainable time-weighting network and base query layers; their learning rate is reduced to $10^{-5}$ for stability. Disease-specific classifiers are trained for $5$ epochs using identical optimization settings.

\subsection{Evaluation Metrics}
For medical code prediction (next visit), we report Acc@$K$ ($K=5,10,20$), macro F1-score, and Recall (details in Appendix~\ref{app:Medical_Code_Prediction}). For disease prediction and clinical benchmark tasks, we use AUROC, AUPRC, and F1-score to assess classification effectiveness and the model's ability to correctly identify disease occurrences while mitigating errors.

\subsection{Baseline Models}

TALE-EHR is compared against several models for EHR modeling. The baseline models include: LSTM \citep{hochreiter1997long}, Retain \citep{retain}, a recurrent attention model that improves interpretability by assigning weights to past visits. RetainEx \citep{kwon2018retainex}, an extension of Retain that enhances interpretability by incorporating additional structured medical information. Dipole \citep{dipole}, a bidirectional recurrent neural network (RNN) that captures temporal dependencies in patient trajectories. HiTANet \citep{hitanet}, a hierarchical Transformer-based model that introduces time-aware attention mechanisms for clinical event prediction. Cehr\_Bert \citep{pang2021cehrbertincorporatingtemporalinformation}, which employs a hybrid strategy for temporal modeling by inserting discrete "Artificial Time Tokens" between visits and concatenating time2vec-generated time and age embeddings with concept embeddings as input to a BERT architecture.

\begin{table*}[htpb!]
    \centering
    \caption{Performance Comparison on Disease Prediction (MIMIC-IV)}
    \label{tab:disease_results} 
    \resizebox{\textwidth}{!}{
    \scriptsize 
    \setlength{\tabcolsep}{3pt} 
    \renewcommand{\arraystretch}{0.9} 
    \begin{tabular}{lccccccccc}
        \toprule

        \multirow{2}{*}{\textbf{Model}} &
        \multicolumn{3}{c}{\textbf{Arteriosclerosis}} & \multicolumn{3}{c}{\textbf{Type 2 Diabetes}} & \multicolumn{3}{c}{\textbf{Hyperlipidemia}}
        \\
        \cmidrule(lr){2-4} \cmidrule(lr){5-7} \cmidrule(lr){8-10}
        & AUROC & AUPRC & F1 & AUROC & AUPRC & F1 & AUROC & AUPRC & F1
        \\
        \midrule
        LSTM     & 0.841±.007 & 0.553±.009 & 0.597±.008 & 0.848±.007 & 0.623±.009 & 0.646±.011 & 0.816±.003 & 0.629±.005 & 0.631±.013 \\
        Retain   & 0.891±.005 & 0.641±.011 & 0.639±.009 & 0.887±.005 & 0.656±.008 & 0.676±.009 & 0.825±.009 & 0.633±.015 & 0.653±.011 \\
        RetainEX & 0.907±.004 & 0.688±.008 & 0.601±.012 & 0.908±.004 & 0.693±.010 & 0.637±.013 & 0.850±.006 & 0.673±.012 & 0.646±.011 \\
        Dipole   & 0.895±.007 & 0.618±.011 & 0.591±.013 & 0.873±.006 & 0.641±.012 & 0.626±.013 & 0.832±.009 & 0.656±.014 & 0.631±.010 \\
        HiTANet  & 0.909±.006 &0.781±.006 & 0.682±.007 & 0.919±.004 & 0.777±.007 & 0.668±.008 & 0.909±.004 & 0.781±.006 & 0.682±.007 \\
        Cehr\_Bert  & 0.912±.004 &0.790±.007 & 0.694±.007 & 0.924±.005 &0.785±.007 & 0.688±.009 & 0.915±.003 & 0.792±.007 & 0.693±.008 \\\midrule
        TALE-EHR & \textbf{0.939$\pm$.005} & \textbf{0.811$\pm$.006} & \textbf{0.744$\pm$.006} & \textbf{0.941$\pm$.004} & \textbf{0.813$\pm$.008} & \textbf{0.743$\pm$.011} & \textbf{0.931$\pm$.004} & \textbf{0.833$\pm$.005} & \textbf{0.741$\pm$.009} \\
        \midrule[\heavyrulewidth]

        \multirow{2}{*}{\textbf{Model}} &
        \multicolumn{3}{c}{\textbf{Depression}} & \multicolumn{3}{c}{\textbf{Gastroesophageal Reflux}} & \multicolumn{3}{c}{\textbf{Acute Kidney Injury}}
        \\
        \cmidrule(lr){2-4} \cmidrule(lr){5-7} \cmidrule(lr){8-10}
        & AUROC & AUPRC & F1 & AUROC & AUPRC & F1 & AUROC & AUPRC & F1
        \\
        \midrule
        LSTM     & 0.822±.009 & 0.529±.007 & 0.532±.011 & 0.782±.007 & 0.611±.010 & 0.488±.010 & 0.871±.006 & 0.581±.013 & 0.509±.011 \\
        Retain   & 0.817±.003 & 0.519±.009 & 0.541±.009 & 0.834±.006 & 0.621±.006 & 0.534±.008 & 0.880±.006 & 0.613±.010 & 0.537±.008 \\
        RetainEX & 0.848±.011 & 0.536±.014 & 0.595±.017 & 0.793±.003 & 0.611±.007 & 0.536±.012 & 0.867±.004 & 0.695±.008 & 0.561±.008 \\
        Dipole   & 0.824±.006 &0.587±.011 & 0.521±.013 & 0.818±.002 & 0.602±.009 & 0.528±.011& 0.897±.005 & 0.624±.012 & 0.608±.007 \\
        HiTANet  & 0.838±.007 &0.610±.009 & 0.545±.011 & 0.856±.005 & 0.644±.006 & 0.508±.009 & 0.938±.007 & 0.628±.007 & 0.597±.013 \\
        Cehr\_Bert  & 0.840±.008 &0.612±.008 & 0.551±.009 & 0.857±.004 & 0.657±.007 & 0.510±.012 & 0.941±.008 & 0.631±.010 & 0.601±.013 \\
        \midrule
        TALE-EHR & \textbf{0.878$\pm$.005} & \textbf{0.637$\pm$.008} & \textbf{0.598$\pm$.010} & \textbf{0.891$\pm$.005} & \textbf{0.698$\pm$.008} & \textbf{0.604$\pm$.007} & \textbf{0.961$\pm$.005} & \textbf{0.670$\pm$.012} & \textbf{0.649$\pm$.011} \\
        \midrule[\heavyrulewidth]

        \multirow{2}{*}{\textbf{Model}} &
        \multicolumn{3}{c}{\textbf{Atrial Fibrillation}} & \multicolumn{3}{c}{\textbf{Heart Failure}} & \multicolumn{3}{c}{\textbf{Hypertension}}
        \\
        \cmidrule(lr){2-4} \cmidrule(lr){5-7} \cmidrule(lr){8-10}
        & AUROC & AUPRC & F1 & AUROC & AUPRC & F1 & AUROC & AUPRC & F1
        \\
        \midrule
        LSTM     & 0.853±.007 & 0.503±.011 & 0.519±.014 & 0.847±.004 & 0.528±.007 & 0.490±.017 & 0.858±.005 & 0.710±.009 & 0.627±.008\\
        Retain   &0.876±.006 & 0.569±.006 & 0.555±.010 &0.870±.003 & 0.541±.006 & 0.538±.008 & 0.886±.005 & 0.727±.012 & 0.663±.009 \\
        RetainEX & 0.867±.007 & 0.506±.009 & 0.504±.010 & 0.874±.004 & 0.543±.007 & 0.533±.013 & 0.919±.008 &0.764±.008 & 0.683±.012 \\
        Dipole   & 0.874±.005 &0.531±.006 & 0.537±.008 &0.902±.006 & 0.566±.010 & 0.570±.015 & 0.879±.003 & 0.756±.006 & 0.658±.013 \\
        HiTANet  & 0.922±.005 & 0.560±.007 & 0.571±.008 & 0.897±.004 & 0.563±.006 & 0.575±.013 &0.909±.006 & 0.782±.009 & 0.681±.014 \\
        Cehr\_Bert  & 0.921±.007 & 0.562±.009 & 0.573±.011 & 0.905±.006 & 0.567±.008 & 0.574±.013 &0.911±.008 & 0.778±.011 & 0.684±.016 \\
        \midrule
        TALE-EHR & \textbf{0.931$\pm$.002} & \textbf{0.630$\pm$.005} & \textbf{0.622$\pm$.004} & \textbf{0.938$\pm$.003} & \textbf{0.661$\pm$.009} & \textbf{0.643$\pm$.012} & \textbf{0.931$\pm$.005} & \textbf{0.812$\pm$.006} & \textbf{0.727$\pm$.010} \\
        \bottomrule
    \end{tabular}
    }
\end{table*}

\section{Results and Discussion} \label{sec:result}

\subsection{Disease and Common Clinical Benchmark Prediction}

 Table~\ref{tab:disease_results} presents classification performance for nine representative diseases, where TALE-EHR achieves the highest scores for all diseases, demonstrating superior  generalization.TALE-EHR’s explicit temporal modeling and history representation enable accurate disease classification across diverse conditions, making it highly applicable for real-world clinical settings. For instance, rapid decay for acute conditions and more gradual decay for chronic ones, as detailed in Appendix~\ref{app:time_aware_2}. 
Table~\ref{tab:mimic_benchmarks} summarizes the results on two common clinical benchmarks. TALE-EHR consistently outperforms all baseline models across AUROC, AUPRC, and F1-score for both benchmark tasks. These results further confirm the effectiveness of TALE-EHR for diverse clinical prediction tasks built upon longitudinal EHR data.

\begin{table}[htpb!]
    \centering
    \caption{Performance on Clinical Benchmark Tasks (MIMIC-IV)}
    \label{tab:mimic_benchmarks}
    \resizebox{\textwidth}{!}{
    \setlength{\tabcolsep}{4pt} 
    \renewcommand{\arraystretch}{0.9}
    \begin{tabular}{lcccccc}
        \toprule
        \multirow{2}{*}{\textbf{Model}} & \multicolumn{3}{c}{\textbf{Hospital 30-day Readmission}} & \multicolumn{3}{c}{\textbf{LOS > 7 days (next visit)}} \\
        \cmidrule(lr){2-4} \cmidrule(lr){5-7}
        & AUROC & AUPRC & F1 & AUROC & AUPRC & F1 \\
        \midrule
        LSTM     & 0.736$\pm$.003 & 0.419$\pm$.002 & 0.442$\pm$.009 & 0.724$\pm$.002 & 0.177$\pm$.005 & 0.227$\pm$.001 \\
        Retain   & 0.710$\pm$.004 & 0.427$\pm$.011 & 0.443$\pm$.004 & 0.740$\pm$.003 & 0.184$\pm$.002 & 0.244$\pm$.011 \\
        RetainEX & 0.734$\pm$.012 & 0.478$\pm$.007 & 0.447$\pm$.003 & 0.743$\pm$.009 & 0.185$\pm$.005 & 0.247$\pm$.017 \\
        Dipole   & 0.739$\pm$.002 & 0.496$\pm$.003 & 0.477$\pm$.005 & 0.745$\pm$.006 & 0.192$\pm$.008 & 0.265$\pm$.012 \\
        HiTANet  & 0.754$\pm$.005 & 0.512$\pm$.009 & 0.522$\pm$.004 & 0.750$\pm$.004 & 0.189$\pm$.003 & 0.261$\pm$.003 \\
        Cehr\_Bert  & 0.756$\pm$.008 & 0.515$\pm$.006 & 0.531$\pm$.005 & 0.751$\pm$.006 & 0.185$\pm$.007 & 0.268$\pm$.009 \\
        \midrule
        TALE-EHR & \textbf{0.762$\pm$.001} & \textbf{0.536$\pm$.004} & \textbf{0.563$\pm$.002} & \textbf{0.759$\pm$.002} & \textbf{0.195$\pm$.002} & \textbf{0.278$\pm$.003} \\
        \bottomrule
    \end{tabular}
    }
\end{table}

\subsection{Generalization to Paediatric Data (PIC)}
To assess generalizability beyond adult populations, TALE-EHR was evaluated on the Paediatric Intensive Care (PIC) dataset \citep{Zeng2020} across four relevant paediatric ICU tasks: Pneumonia, Heart Malformations, in-hospital Mortality, and ICU length of stay >7 days. 
As shown in Table~\ref{tab:pic_generalization}, TALE-EHR consistently outperformed baseline models across all four tasks, achieving the highest AUROC, AUPRC, and F1 scores. This strong performance on PIC data highlights TALE-EHR's robustness and ability to generalize across different clinical environments, patient populations (paediatric vs. adult), and prediction tasks, underscoring its broad applicability.

\begin{table}[htpb!] 
    \centering
    \caption{Performance on PIC Dataset } 
    \label{tab:pic_generalization} 
    \resizebox{\textwidth}{!}{
    \small 
    \setlength{\tabcolsep}{4pt} 
    \renewcommand{\arraystretch}{0.9} 
    \begin{tabular}{@{}lcccccc@{}} 
        \toprule
        & \multicolumn{3}{c}{\textbf{Pneumonia Prediction}} & \multicolumn{3}{c}{\textbf{Heart Malformations Prediction}} \\
        \cmidrule(lr){2-4} \cmidrule(lr){5-7} 
        \textbf{Model} & AUROC & AUPRC & F1 & AUROC & AUPRC & F1 \\ 
        \midrule
        LSTM     & 0.885$\pm$.001 & 0.463$\pm$.003 & 0.550$\pm$.004 & 0.825$\pm$.002 & 0.206$\pm$.005 & 0.329$\pm$.007 \\
        Retain   & 0.912$\pm$.002 & 0.538$\pm$.003 & 0.602$\pm$.005 & 0.856$\pm$.003 & 0.236$\pm$.004 & 0.335$\pm$.004 \\
        RetainEX & 0.902$\pm$.006 & 0.537$\pm$.009 & 0.582$\pm$.010 & 0.867$\pm$.005 & 0.267$\pm$.004 & 0.342$\pm$.007 \\
        Dipole   & 0.920$\pm$.005 & 0.559$\pm$.007 & 0.598$\pm$.004 & 0.866$\pm$.005 & 0.253$\pm$.006 & 0.342$\pm$.005 \\
        HiTANet  & 0.929$\pm$.003 & 0.568$\pm$.007 & 0.621$\pm$.009 & 0.886$\pm$.002 & 0.310$\pm$.002 & 0.382$\pm$.004 \\
        Cehr\_Bert  & 0.933$\pm$.004 & 0.573$\pm$.011 & 0.628$\pm$.006 & 0.884$\pm$.006 & 0.313$\pm$.005 & 0.384$\pm$.006 \\
        \midrule
        TALE-EHR & \textbf{0.945$\pm$.002} & \textbf{0.613$\pm$.003} & \textbf{0.651$\pm$.006} & \textbf{0.906$\pm$.003} & \textbf{0.340$\pm$.005} & \textbf{0.409$\pm$.005} \\
        \midrule[\heavyrulewidth] 
        & \multicolumn{3}{c}{\textbf{Mortality Prediction}} & \multicolumn{3}{c}{\textbf{ICU LOS > 7 days}} \\
         \cmidrule(lr){2-4} \cmidrule(lr){5-7}
         & AUROC & AUPRC & F1 & AUROC & AUPRC & F1 \\ 
        \midrule
        LSTM     & 0.894$\pm$.004 & 0.373$\pm$.004 & 0.448$\pm$.006 & 0.841$\pm$.003 & 0.679$\pm$.002 & 0.664$\pm$.005 \\
        Retain   & 0.904$\pm$.003 & 0.422$\pm$.005 & 0.450$\pm$.008 & 0.851$\pm$.002 & 0.700$\pm$.005 & 0.665$\pm$.006 \\
        RetainEX & 0.912$\pm$.009 & 0.423$\pm$.011 & 0.451$\pm$.006 & 0.859$\pm$.006 & 0.715$\pm$.007 & 0.683$\pm$.008 \\
        Dipole   & 0.913$\pm$.006 & 0.421$\pm$.009 & 0.456$\pm$.005 & 0.863$\pm$.001 & 0.729$\pm$.004 & 0.685$\pm$.002 \\
        HiTANet  & 0.921$\pm$.004 & 0.429$\pm$.005 & 0.464$\pm$.003 & 0.874$\pm$.002 & 0.747$\pm$.003 & 0.705$\pm$.004 \\
        Cehr\_Bert  & 0.923$\pm$.003 & 0.431$\pm$.005 & 0.467$\pm$.004 & 0.877$\pm$.004 & 0.751$\pm$.006 & 0.713$\pm$.009 \\
        \midrule
        TALE-EHR & \textbf{0.934$\pm$.002} & \textbf{0.443$\pm$.005} & \textbf{0.475$\pm$.004} & \textbf{0.897$\pm$.001} & \textbf{0.784$\pm$.002} & \textbf{0.741$\pm$.004} \\
        \bottomrule
    \end{tabular}
    }
\end{table}
\section{Ablation Study}
\label{sec:ablation}
\begin{table*}[tp]
    \centering
    \small
    \setlength{\tabcolsep}{3pt}
    \renewcommand{\arraystretch}{0.9}
    \caption{Ablation Study Results on Disease Prediction evaluated using AUROC.(MIMIC-IV) }
    \label{tab:disease_results_ablation}
    \resizebox{\textwidth}{!}{
    \begin{tabular}{lcccccc}
        \toprule
        \textbf{Disease} & \textbf{Poly=5} & \textbf{MLP} & \textbf{Poly=10} & \textbf{Piecewise} & \textbf{W/O time}& \textbf{Random Embed} \\
 
        \midrule
        Arteriosclerosis  & \textbf{0.939±.005} & 0.937±.006 & 
        0.934±.004 & 0.922±.007 & 0.911±.008 &
        0.927±.010\\
        Type 2 Diabetes & 0.941±.004 & \textbf{0.942±.006} & 
        0.937±.006 & 0.927±.006 & 0.919±.008 &
        0.931±.007\\
        Hyperlipidemia & \textbf{0.931±.004} & 0.928±.005 & 
        0.925±.006 & 0.919±.006 & 0.912±.003 &
        0.922±.006\\
        Depression  & \textbf{0.878±.005 } & 0.875±.007  & 
        0.870±.006  & 0.859±.005  & 0.844±.009  &
        0.863±.009\\
        Gastroesophageal Reflux & \textbf{0.891±.005 } & 0.889±.004  & 
        0.882±.008  & 0.869±.003  & 0.852±.005  &
        0.873±.012\\
        Acute Kidney Injury & 0.961±.005 & \textbf{0.964±.005} & 0.957±.004&0.948±.007 & 0.938±.003&0.956±.010\\
        Atrial Fibrillation  & \textbf{0.931±.002} & 0.930±.004 & 
        0.927±.006 & 0.923±.005 & 0.914±.006 &
        0.925±.013\\
        Heart Failure  & \textbf{0.938±.003} & 0.936±.005 & 
        0.930±.004 & 0.921±.003 & 0.902±.008 &
        0.931±.008\\
        Hyperlipidemia & 0.931±.005 & \textbf{0.932±.004} & 
        0.927±.006 & 0.919±.005 & 0.908±.004 &
        0.925±.009\\
        \bottomrule
    \end{tabular}
    }
\end{table*}

The ablation study systematically evaluates key design choices within TALE-EHR, focusing on: (1) the temporal encoding strategy, (2) the impact of the time-aware mechanism, and (3) the contribution of the embedding initialization approach. Through these experiments, we aim to validate our model design choices and understand their individual contributions to overall disease prediction performance. We consider the following model variants, with results presented in Table~\ref{tab:disease_results_ablation}:
\begin{itemize}
\item \textbf{Polynomial Basis (Order 5) [Ours, Poly=5]}: Our main TALE-EHR model, learns the time-aware representations within the attention mechanism using a polynomial function of degree 5.
\item \textbf{MLP for Time-Awareness [MLP]}: Replaces the polynomial basis function with an MLP to learn the time-aware representations within the attention mechanism.
\item \textbf{Polynomial Basis (Order 10) [Poly=10]}: Extends the polynomial function to degree 10 to assess the impact of increased temporal expressiveness.
\item \textbf{Piecewise Function for Time Encoding [Piecewise]}: Employs a predefined set of 7 discrete time intervals (1-7 days, 7-30 days, 30-90 days, 90-180 days, 180-365 days, 365-720 days, and >720 days) to represent time, implemented as indicator functions.
\item \textbf{Without Time-Aware Mechanism [W/O time]}: Removes the explicit time-aware component from TALE-EHR, forcing the model to rely solely on sequence order without explicit temporal distance modeling. This variant still uses LLM-derived embeddings.
\item \textbf{Random Initialized Embedding [Random Embed]}: Replaces LLM-derived embeddings with random ones, retaining Poly=5 time-awareness, to isolate pre-trained knowledge contribution.
\end{itemize}
Table~\ref{tab:disease_results_ablation} presents AUROC scores for these variants. Our TALE-EHR with Polynomial Basis (Order 5)  demonstrates strong performance . An MLP-based time-awareness variant yields comparable efficacy, However, Poly=5 was selected for its balance of performance and structured efficiency.
Increasing the polynomial order to 10  slightly degrades performance, likely due to overfitting or increased sensitivity to minor temporal variations. 
A Piecewise Function approach shows a more noticeable decline, suggesting predefined intervals lack flexibility for nuanced temporal dependencies.
The critical role of the time-aware mechanism is highlighted by the "W/O time aware" variant. Lacking explicit temporal modeling, its performance drops significantly across all diseases , underscoring that explicitly modeling temporal relationships is paramount for effective clinical event sequence modeling.

Finally, comparing Poly=5 with LLM embeddings to a variant with Random Initialized Embeddings (which retains Poly=5 time encoding) assesses the impact of pre-trained knowledge. While the random embedding model performs respectably, it consistently underperforms its LLM-equipped counterpart and exhibits increased result variance. This confirms LLM embeddings enhance predictive accuracy and stability. Notably, the performance gains from the time-aware mechanism itself are more substantial than those lost by switching from LLM to random embeddings. This suggests that while LLM embeddings offer a significant advantage, the sophisticated time-aware architecture is the primary driver of TALE-EHR's strong performance.

These findings validate our design choices: TALE-EHR, with polynomial time representations (Order 5) and augmented by pre-trained LLM embeddings, offers a robust, high-performing EHR modeling framework. The study emphasizes the foundational importance of the explicit time-aware mechanism, with LLM embeddings serving as a powerful performance and stability enhancer.

\section{Conclusion}
We introduced TALE-EHR, a novel framework synergistically integrating sophisticated time-aware attention with rich semantic understanding from pre-trained LLM embeddings, significantly enhancing EHR modeling. Extensive experiments show TALE-EHR effectively captures intricate clinical dependencies, consistently outperforming state-of-the-art baselines in medical code and diverse disease prediction tasks. Evaluations on MIMIC-IV and PIC datasets confirm its superior predictive accuracy, especially for complex conditions requiring nuanced interpretation of long-term patient context. Comprehensive ablation studies rigorously underscore the indispensable role of its explicit, continuous temporal modeling. By fusing advanced temporal modeling with deep pre-trained medical knowledge, TALE-EHR advances EHR analytics, paving the way for more reliable real-world clinical decision-making support. Future work includes extending this framework to incorporate multimodal data, like clinical notes and imaging, for a more holistic patient understanding.

\newpage
\bibliography{ref}
\bibliographystyle{chicago}
\clearpage
\onecolumn
\newpage
\appendix

\section*{Appendix}
\renewcommand{\thesubsection}{\Alph{subsection}}

\section{Training Algorithm}
\label{app:algorithm}

Algorithm~\ref{app:training} summarizes the training procedure outlined in Section~\ref{sec:method}, consisting of pre-training on medical event sequences and fine-tuning for disease prediction. The pre-training phase learns structured temporal and semantic representations of clinical events, while fine-tuning optimizes the model for disease-specific predictions.

\begin{algorithm}[H]
\caption{Time-Aware Language Encoder for EHR (TALE-EHR)}
\label{app:training}
\begin{algorithmic}[1]
    \Statex \textbf{Input}: Patient histories $\{\mathcal{H}^{(i)}\}$, medical code set $\mathcal{C}$
    \Statex \textbf{Output}: Trained model parameters $\theta$
    \Statex
    \State \textcolor{blue}{\textit{// Pre-training Phase}}
    \State Initialize model parameters $\theta$ 
    \For{each medical code $c \in \mathcal{C}$}
        \State $x_c \gets \text{GetDescription}(c)$ \Comment{Map code to standardized description}
        \State $\mathbf{v}_c \gets \text{BGE}(x_c)$ \Comment{Generate embeddings from pretrain LLM model}
    \EndFor
    \For{epoch = 1 to $N_{\text{pretrain}}$}
        \For{each batch $\mathcal{B}$}
            \For{each patient history $\mathcal{H}_t^{(i)}$ in $\mathcal{B}$}
                \State Compute code-level representations using time-aware attention
                \State Generate unified history representation $\mathbf{h}_t^{(i)}$ for current time $t$
                \State Compute temporal and code prediction losses
            \EndFor
            \State Update model parameters using combined loss
        \EndFor
    \EndFor
    \Statex 
    \State \textcolor{blue}{\textit{// Fine-tuning Phase for Disease Prediction}}
    \State Initialize task-specific parameters
    \State Configure learning rates for pre-trained and new parameters
    \For{epoch = 1 to $N_{\text{finetune}}$}
        \For{each batch $\mathcal{B}$}
            \State \textcolor{blue}{\textit{// Disease Prediction}}
            \State Extract histories before disease onset time
            \State Generate pre-onset unified representations and compute disease loss
            \State Update model parameters
        \EndFor
    \EndFor
    \State \textbf{return} Optimized model parameters $\theta$
\end{algorithmic}
\end{algorithm}

\section{Code Description Standardization}
\label{sec:code_description}

After basic preprocessing, our code set primarily consists of six types of codes: ICD, RxNorm, Phecode, CCS, DRG, and MIMIC-IV local codes. We processed these codes and obtained their descriptions as follows. For ICD (International Classification of Diseases) codes, we used the descriptions provided in MIMIC-IV, which contains official WHO (\url{https://www.who.int/standards/classifications/classification-of-diseases}) descriptions that provide detailed explanations of each diagnosis. For RxNorm medication codes, we employed the standardized drug descriptions from the National Library of Medicine (NLM) (\url{https://www.nlm.nih.gov/research/umls/rxnorm}). For Phecode (Phenotype Code), we utilized the standardized definitions and descriptions from the PheWAS Catalog (\url{https://phewascatalog.org}) maintained by Vanderbilt University, which maps clinical phenotypes to ICD codes. For CCS (Clinical Classifications Software) codes, we incorporated the descriptions provided by HCUP (Healthcare Cost and Utilization Project) (\url{https://hcup-us.ahrq.gov/toolssoftware/ccs/ccs.jsp}) that categorize medical procedures and services into clinically meaningful groups. For DRG (Diagnosis Related Groups) codes, the EHR contains two classification systems: APR-DRG by 3M (\url{https://www.solventum.com/en-us/home/h/f/b5005024009/}) which incorporates severity and mortality risk scores, and HCFA-DRG by CMS (\url{https://www.cms.gov/medicare/payment/prospective-payment-systems/acute-inpatient-pps/ms-drg-classifications-and-software}) for Medicare reimbursement. We mapped these codes to descriptions using files provided in MIMIC-IV. Lastly, for local institution-specific codes, we use the institution's coding manual descriptions.

\section{Data Integration and Code Roll-up}
\label{app:roll-up}
The data (\href{https://mimic.mit.edu/docs/}{https://mimic.mit.edu/docs/}) contains the EHR of $364,627$ patients, out of them $94,458$ were in the ICU and out of them $205,504$ have attended Emergency department(ED). Patients who have been in the ICU tend to relate to much more codes and occurrences than other patients.\par
These data comes from \textbf{hospital tables}: diagnoses\_icd, drgcodes, hcpcsevents, labevents, prescriptions, procedures\_icd, \textbf{ICU tables}: chartevents, inputevents, outputevents, procedureevents and \textbf{ED tables}: medrecon, pyxis, diagnosis.\par
Some codes from these tables are aggregated into high-level codes. We convert procedure codes, including HCPC, ICD-9, ICD-10-PCS, to CCS codes. Diagnosis codes, including ICD-9 and ICD-10-CM, are grouped to PheCodes. NDC codes are aggregated to RxNorm CUIs. The statistics of these aggregated codes are shown in Table \ref{tab:aggregation}.\par

\begin{table}[h]
\centering
\caption{Statistics of aggregated code}
\label{tab:aggregation}
\begin{tabular}{cccc}
\hline
Original Code & Source Table    & Number of original code & Target code \\ \hline
ICD9/ICD10PCS & procedures\_icd & 14,911                  & CCS         \\
HCPCS         & hcpcsevents     & 2,366                   & CCS         \\
ICD9/ICD10CM  & diagnoses\_icd/diagnosis  & 29,933                  & PheCode     \\
NDC           & prescriptions/medrecon/pyxis   & 12,243                  & RxNorm      \\ \hline
\end{tabular}
\end{table} 

\begin{table}[h]
\centering
\caption{Statistics of final dataset}
\label{tab:final-statistics}
\resizebox{\textwidth}{!}{
\begin{tabular}{ccccccc}
\hline
Code        & Source Table     & Number of codes & Total freq  & Mean freq & Median freq & Max freq  \\ \hline
Chart code  & chartevents      & 1,891            & 414,200,715 & 219,037   & 16,078     & 12,869,833 \\
Input code  & inputevents      & 327             & 10,953,713   & 33,497    & 2,300       & 1,578,656 \\
Output code & outputevents     & 66              & 5,359,187   & 81,199    & 8,782      & 3,599,702 \\
Proc code   & procedureevents  & 159               & 808,706     & 5,086    & 501       & 101,294    \\
Lab code    & labevents        & 734             & 158,084,819 & 215,374   & 5,779      & 4,331,615 \\
ICD10     & diagnoses\_icd/diagnosis  & 1830              & 242,533      & 132     & 4      & 33,113    \\
ICD9        & diagnoses\_icd/diagnosis   & 954     & 253,908     & 266     & 10       & 12,600    \\
PheCode     & diagnoses\_icd/diagnosis   & 1,793       & 6,766,921   & 3,774     & 557       & 237,879   \\
DRG         & drgcodes         & 1,087             & 761,856     & 700    & 267       & 12,633    \\
CCS         & hcpcs/proc\_icd* & 234             & 1,041,681     & 4,451     & 1.587       & 157,740  \\
RxNorm      & prescriptions/medrecon    & 3,157               & 20,890,607  & 6,617    & 30      & 1,048,814   \\
Total       & -                & 12,232            & 619,364,646 & 50,634   & 222       & 12,869,833 \\ \hline
\end{tabular}
}
*CCS codes are derived from hcpcsevents and procedures\_icd.\par
\end{table}
 Some ICD codes from diagnoses\_icd with high frequency cannot be grouped to PheCode by the mapping table we use, and we still keep them in the dataset. Eventually, we split some codes from Chartevents into several distinguishable codes, because they occur with only a few unique values. Statistics of code occurrences of the final dataset are shown in Table \ref{tab:final-statistics}.\par

\section{Temporal Weight Analysis}
\label{app:time_aware_2}
\begin{figure}[t]
    \centering
    \includegraphics[width=\linewidth]{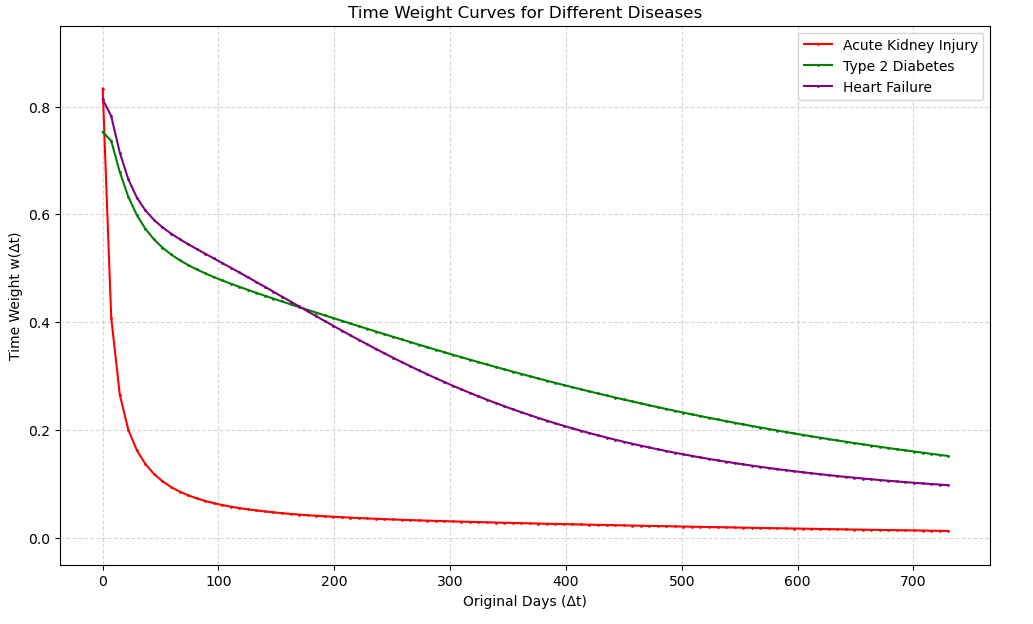}
    \caption{Learned temporal weighting function $w(\Delta t)$ for downstream disease prediction tasks. The x-axis represents the time interval between clinical events (in days), and the y-axis shows the attention weight. The curve demonstrates how the model modulates attention across different temporal scales, with a gradual decay indicating that both recent and historical events contribute to disease prediction, though with decreasing importance over time.}
    \label{fig:time_weight.png}
\end{figure}

To gain deeper insights into how TALE-EHR models temporal influence, we analyze the learned temporal weighting function $w(\Delta t)$ for different downstream disease prediction tasks. Figure~\ref{fig:time_weight.png} visualizes these learned weighting functions, revealing distinct characteristics based on the nature of the diseases.

First, the analysis highlights Disease-Specific Decay Rates, where the temporal weight decay varies significantly across different diseases. For instance, the curve associated with Acute Kidney Injury (an acute condition, represented by the red line in Figure~\ref{fig:time_weight.png}) exhibits rapid decay. This indicates that very recent clinical events hold substantially higher importance for predicting acute illnesses, and the influence of past events diminishes quickly.

Second, a Gradual Decay for Chronic Conditions is observed. In contrast to acute conditions, the weighting functions for chronic conditions like Type 2 Diabetes (green line in Figure~\ref{fig:time_weight.png}) and Heart Failure (purple line in Figure~\ref{fig:time_weight.png}) show a much more gradual decay. This suggests that for chronic diseases, historical events, even those occurring further in the past, retain a more significant and prolonged influence, reflecting the model's recognition of their long-term, cumulative impact.

Finally, a Long-Tail Effect for Capturing Distant Influences is notable. Even for diseases with slower decay patterns, the weights do not abruptly drop to zero but instead exhibit a long-tail effect. This implies that very distant events can still maintain small but non-zero weights, allowing the model to capture rare but potentially significant long-term clinical patterns or risk factors that might be relevant much later.

These observations collectively demonstrate that TALE-EHR's time-aware mechanism capably learns and adapts disease-specific temporal dependencies. It effectively captures the varying importance of events over time, differentiating between the acute, rapidly evolving nature of some conditions and the chronic, long-term influence patterns of others. This adaptability is crucial for building accurate and clinically relevant predictive models from EHR data.

\section{Medical Code Prediction}
\label{app:Medical_Code_Prediction}
\begin{table}[htpb!]
    \centering
    \setlength{\tabcolsep}{3pt}
    \renewcommand{\arraystretch}{0.9} 
    \caption{Performance in Medical Code Prediction(MIMIC-IV)}
    \label{model-comparison}
    \begin{center}
    \begin{tabular}{lccccc}
    \toprule
    \textbf{Model} & \textbf{Acc@5} & \textbf{Acc@10} & \textbf{Acc@20} & \textbf{F1} & \textbf{Recall}\\
    \midrule
    
    LSTM &  0.829±.004&  0.824±.003&  0.803±.004&  0.718±.006& 0.627±.009\\
    Retain  & 0.869±.002 & 0.861±.004 & 0.842±.007  &  0.755±.005& 0.667±.023\\
    RetainEx  & 0.882±.003 & 0.879±.006 & 0.830±.010  &  0.844±.008& 0.615±.015\\
    Dipole  & 0.869±.004 & 0.862±.004 & 0.834±.004 & 0.730±.004 & 0.673±.011\\
    HiTANet & 0.861±.004 & 0.817±.008 & 0.787±.009  &  0.684±.007& 0.626±.013\\
    Cehr\_Bert  & 0.881±.003 & 0.878±.006 & 0.848±.009  & 0.842±.008 & 0.668±.010\\\midrule
    TALE-EHR   & \textbf{0.902±.003}   &  \textbf{0.896±.005} & \textbf{0.862±.006} & \textbf{0.850±.004}  & \textbf{0.675±.007} \\
    \bottomrule
    \end{tabular}
    \end{center}
    \vskip -0.1in
\end{table}
Table~\ref{model-comparison} presents the performance of various models on the medical code prediction task using the MIMIC-IV dataset. TALE-EHR consistently and significantly outperforms all baseline models across all evaluated metrics, achieving top scores in Acc@5 (0.902), Acc@10 (0.896), Acc@20 (0.862), F1-score (0.850), and Recall (0.675). This demonstrates its superior capability in modeling complex medical code dependencies and accurately predicting a comprehensive set of future codes.
Among the baseline models, Cehr\_Bert consistently demonstrates robust performance across the Acc@K metrics (e.g., Acc@5: 0.881, Acc@20: 0.848), making it one of the strongest baselines. This may be attributed to BERT's inherent masked language modeling pre-training, which often aids in predictive tasks. However, TALE-EHR still surpasses Cehr\_Bert in all aspects, particularly in F1-score and Recall.
RetainEx exhibits a strong start, achieving high scores in Acc@5 (0.882) and a competitive F1-score (0.844), second only to TALE-EHR in the latter. This suggests effectiveness in short-term predictions. However, its performance notably degrades for longer-range predictions (Acc@20 drops to 0.830) and, critically, it records the lowest Recall (0.615) among all compared models. This indicates a significant weakness in capturing a comprehensive set of relevant future codes, especially as the prediction horizon extends. In contrast, TALE-EHR maintains superior accuracy and recall even for larger K.
Other models like Retain and Dipole show competitive scores in specific metrics (e.g., Dipole's Recall at 0.673 is strong among baselines), but none match the consistent, across-the-board superiority of TALE-EHR. HiTANet generally underperforms in this task.
The leading F1-score (0.850) and Recall (0.675) achieved by TALE-EHR are particularly noteworthy. A high Recall is crucial in medical applications as it signifies a lower rate of false negatives, meaning fewer actual subsequent medical codes are missed. The high F1-score indicates an excellent balance between precision and recall, making TALE-EHR robust, especially when dealing with the often imbalanced nature of medical code distributions.
By effectively integrating its Transformer-based point process framework, TALE-EHR adeptly captures both short- and long-term temporal dependencies inherent in patient visit sequences. This capability translates directly into its superior predictive accuracy across various horizons (K values) and its ability to identify a more complete set of future codes. Consequently, TALE-EHR emerges as a particularly well-suited model for clinical decision-making, where minimizing missed diagnoses and achieving overall predictive accuracy are critical.

\section{Visualization of Learned Disease Representations}
\label{app:umap_visualization}

To further investigate the quality of the learned patient history representations $\mathbf{h}_t^{(i)}$, we visualize them using UMAP for several representative diseases from the MIMIC-IV dataset. These visualizations project the high-dimensional history representations into a 2D space, allowing for an intuitive assessment of their separability for disease classification. We compare the representations learned by our TALE-EHR model against those learned by a standard LSTM baseline. In these visualizations, red points denote patients who eventually developed the specific disease (positive cases), while blue points represent patients who did not (negative cases).

\begin{figure}[htpb!]
    \centering
    \includegraphics[width=0.65\textwidth]{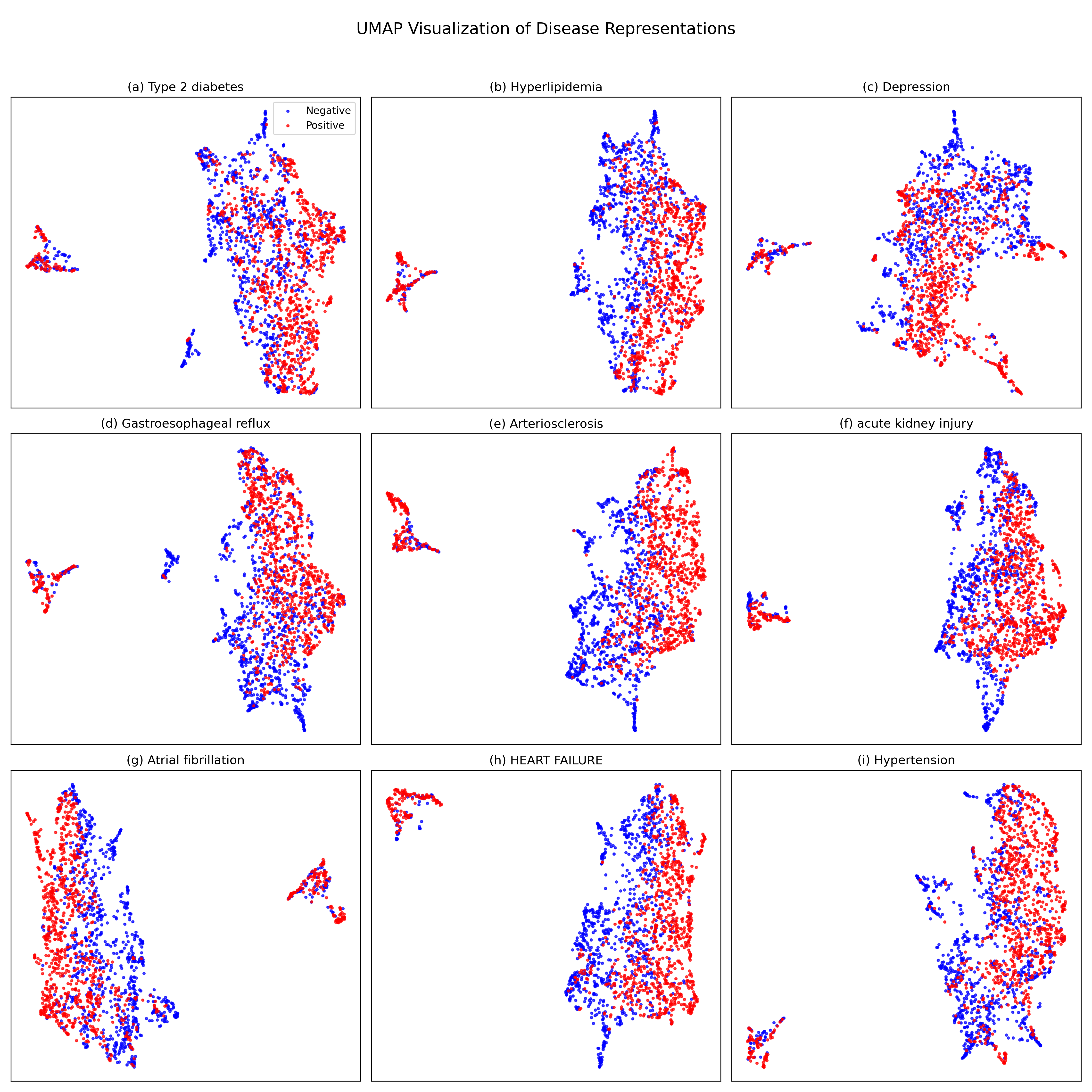} 
    \caption{UMAP visualization of disease representations learned by TALE-EHR for nine diseases. Red points indicate positive cases (disease present), and blue points indicate negative cases (disease absent). Clearer separation between red and blue clusters suggests more discriminative representations.}
    \label{fig:umap_tale_ehr}
\end{figure}

Figure~\ref{fig:umap_tale_ehr} displays the UMAP visualizations for disease representations learned by TALE-EHR. Across various diseases, such as Type 2 Diabetes, Arteriosclerosis, and Heart Failure, we observe a discernible separation between the clusters of positive and negative samples. While some overlap is expected due to the inherent complexity and comorbidity in EHR data, the overall clustering trend indicates that TALE-EHR learns representations that are informative for distinguishing between disease outcomes.

\begin{figure}[htpb!]
    \centering
    \includegraphics[width=0.65\textwidth]{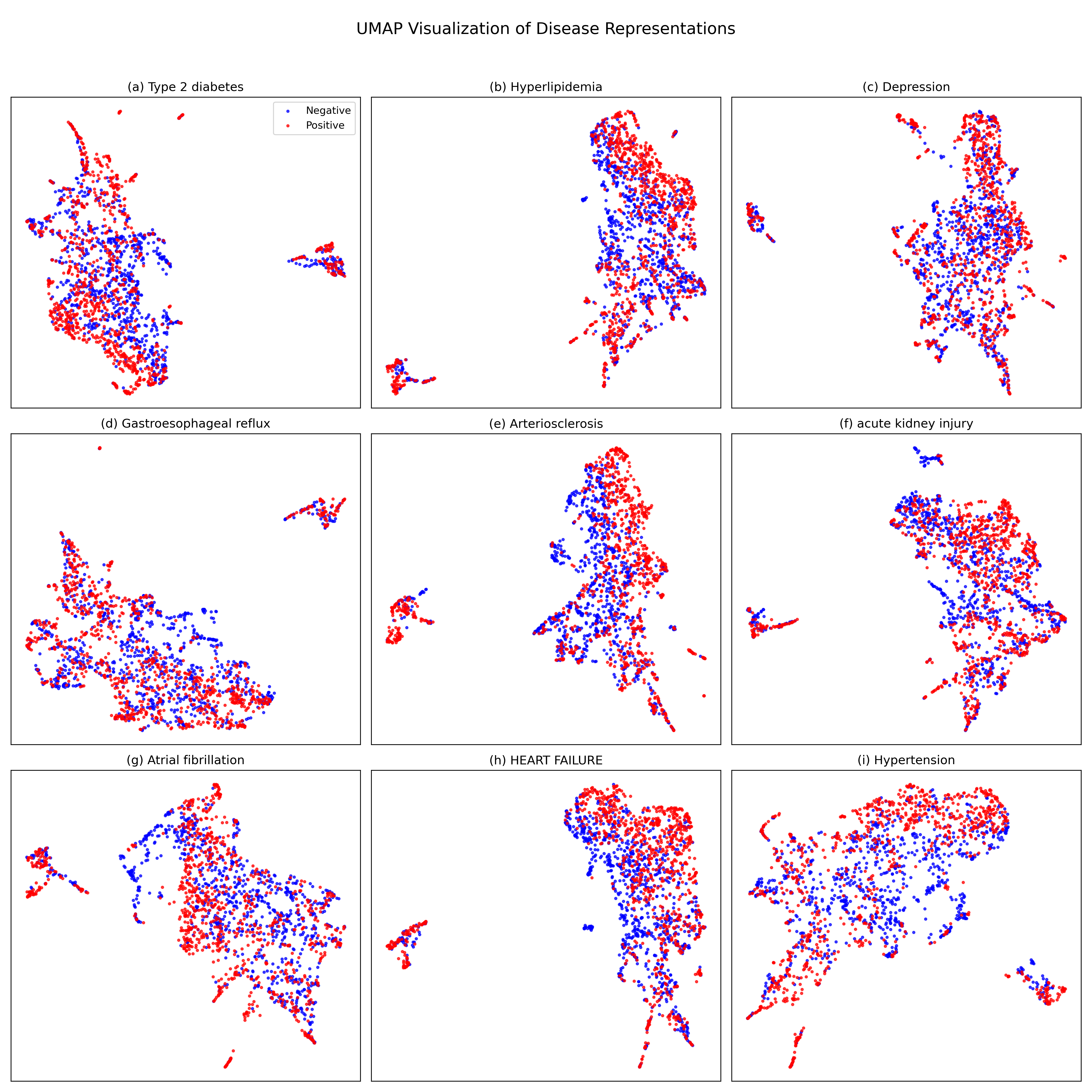} 
    \caption{UMAP visualization of disease representations learned by LSTM for the same nine diseases. Comparison with Figure~\ref{fig:umap_tale_ehr} shows generally less distinct separation between positive (red) and negative (blue) cases for LSTM.}
    \label{fig:umap_lstm}
\end{figure}

In contrast, Figure~\ref{fig:umap_lstm} shows the UMAP visualizations for representations learned by the LSTM baseline. For many of the same diseases, the separation between positive and negative samples appears less distinct compared to TALE-EHR. The clusters often exhibit more significant overlap, and the overall structure seems less organized with respect to the disease labels.

This qualitative comparison suggests that TALE-EHR's architecture, particularly its time-aware attention mechanism and integration of rich semantic embeddings, leads to more discriminative patient history representations. The enhanced separability observed in the TALE-EHR visualizations aligns with its superior quantitative performance in disease prediction tasks (as shown in Table~\ref{tab:disease_results}), providing further evidence of its effectiveness in capturing clinically relevant patterns from complex longitudinal EHR data.

\newpage
\section*{Impact Statement}

TALE-EHR represents an advancement in EHR modeling by integrating a novel time-aware attention mechanism with strong semantic representations derived from LLM embeddings of standardized code descriptions. This combination significantly improves predictive accuracy for critical clinical tasks such as medical code forecasting and disease progression analysis, as demonstrated by robust performance across diverse datasets (MIMIC-IV and PIC) and standard clinical benchmarks. By more accurately capturing both the complex temporal dynamics and the semantic nuances within patient records, the framework enhances the potential for developing reliable clinical decision support tools. The use of LLM-derived embeddings promotes semantic consistency, contributing to the model's ability to handle varied clinical coding practices effectively. Ultimately, this work could facilitate a deeper understanding of patient trajectories and lead to more dependable outcome predictions, benefiting clinical research and potentially improving patient care pathways.

\end{document}